\documentclass[format=acmsmall, review=false]{acmart}
\usepackage{acm-ec-20}
\usepackage{amsthm}
\setcitestyle{acmnumeric}
\pdfoutput=1

\usepackage{amsmath}
\usepackage{graphicx}
\usepackage[utf8]{inputenc}
\usepackage{algorithm}
\usepackage[noend]{algpseudocode}
\usepackage{parskip}

\newcommand{\norm}[1]{\left\lVert#1\right\rVert}

\DeclareMathOperator*{\argmin}{arg\,min}
\DeclareMathOperator*{\argmax}{arg\,max}

\title{Off-Policy Optimization of Portfolio Allocation Policies under Constraints}
\author{Nymisha Bandi}
\affiliation{nymisha.bandi@gmail.com}
\email{nymisha.bandi@gmail.com}
\author{Theja Tulabandhula}
\affiliation{tt@theja.org}
\email{tt\@theja.org}

\begin{abstract}
The dynamic portfolio optimization problem in finance frequently requires learning policies that adhere to various constraints, driven by investor preferences and risk. We motivate this problem of finding an allocation policy within a sequential decision making framework and study the effects of: (a) using data collected under previously employed policies, which may be sub-optimal and constraint-violating, and (b) imposing desired constraints while computing near-optimal policies with this data. Our framework relies on solving a minimax objective, where one player evaluates policies via off-policy estimators, and the opponent uses an online learning strategy to control constraint violations. We extensively investigate various choices for off-policy estimation and their corresponding optimization sub-routines, and quantify their impact on computing constraint-aware allocation policies. Our study shows promising results for constructing such policies when back-tested on historical equities data, under various regimes of operation, dimensionality and constraints.
\end{abstract}

\begin{document}
\sloppy
\maketitle

\section{Introduction}
Portfolio optimization in finance is a well studied problem where our objective is to maximize the returns on assets, typically stocks, by appropriately investing in them over time. There are two key steps one needs to address in portfolio optimization. The first is to determine which stocks to invest in, and the second is to figure out how much to invest in each stock to maximize the returns. In \cite{fu2018machine}, the authors discuss different machine learning and deep learning methods for stock selection. Similarly, \cite{perrin2019machine} and \cite{yu2019model} study the redistribution of wealth across different stocks/investments. Our goal is to combine these two problems into a single stock selection and allocation problem that works with offline data, wherein we also critically focus on meeting \emph{investor preferences} (which might be absent while collecting said offline data). These preferences could be related to risk or other considerations such as minimizing exposure to stocks related to certain industries. In particular, we optimize the weights of different stocks in the portfolio and constantly redistribute these weights to ensure the maximum returns over a period of time while imposing exogenous constraints.

There is a pool of behavioural data based on previous trades that is typically available, which can be leveraged to decide the portfolio redistribution weights at any given time step as market conditions evolve. For instance, \cite{yu2019model} proposes an imitation learning model that produces a portfolio strategy which mimics an expert's behaviour at any given market state. Noticeably, it does not try to improve upon the expert's policy further. When an expert makes a decision to buy or sell a stock, they look at various factors such as the economy, company's management decisions, current investors, dividend yields etc. Each expert is biased towards different metrics when they make these decisions. It would be very useful to observe trading behavior of multiple experts and synthesize a single trading strategy that attains a rate of return which is ideally more than any single trader's returns, while at the same time adhering to new desirable constraints.

Our solution methodology is as follows. We use a batch policy learning method that works under multiple constraints to solve for a return maximizing portfolio strategy using previously collected data. Batch learning implies that the entire data (returns and prices of the concerned stocks and related metadata) is available at the beginning of the learning process, unlike online learning (or other online sequential decision making settings). In this sense, the portfolio optimization resembles a supervised learning problem (more details in Section~\ref{sec:approach}). This approach is motivated by the work of \cite{le2019batch}, where they propose a constrained batch learning algorithm for finding optimal policies in finite action spaces. Since portfolio weights are continuous, we modifying the underlying algorithmic approach for continuous action spaces. Along the way, we also rigorously assess the suitability of nonlinear policy function approximators (notably, using popular neural network architectures) as well as off-policy estimation techniques, both of which are critical in the learning of a near-optimal policy using previously collected data.

This paper is split into three parts. The first discusses the problem mathematically and how the behavioural data is collected (Section~\ref{sec:setting}), the second part discusses a constrained batch learning algorithmic template for continuous action spaces (Section~\ref{sec:approach}), and the third part discusses the results obtained using the proposed portfolio optimization strategy and their comparison with common baselines, allowing us to illustrate their strengths and weaknesses  (Section~\ref{sec:experiments}).

\section{Problem Setting}\label{sec:setting}
We model our problem as a Markov Decision Process (MDP), where the agent (trader) interacts with an environment (market). The MDP is represented using a tuple  $(X,A,r,g,P,\gamma)$, where $X \subset \mathbb{R}^n$ is a continuous state space; $A \subset \mathbb{R}^m$ is a continuous action space representing weights on the stocks in the portfolio; $r(x,a)$ is the reward  (profit) function upon taking action $a$ in $A$; $g(x,a)$ is the constraint cost function (e.g., risk) at state $x$ when we take action $a$; $P$ is the transition model that maps the state/action pairs to next state; and $\gamma \in (0,1)$ is the discount factor. 

We now briefly introduce a few financial terms relevant to portfolio optimization and the associated notation.  A stock is defined as a type of security that signifies proportionate ownership in the issuing corporation. This entitles the stockholder to that proportion of the corporation's assets and earnings. The price of the stock $i$ at the opening of the market at time $t$ is called the opening price, $p_{i,t}^{open}$ and at the close of the market is the closing price, $p_{i,t}^{close}$. We use relative price, $p_{i,t}=\frac{p_{i,t}^{open}}{p_{i,t-1}^{close}}$ to make the buy and sell decisions for each stock\cite{jiang2017deep}. The vector of relative prices at time $t$ for all the stocks in the portfolio is denoted as 
\begin{equation}
v_t = [1,p_{1,t}^{relative},\dots ,p_{N,t}^{relative}], 
\end{equation}
where $p_{0,t}^{relative} = 1$, since it represents the relative price of the capital invested. 
The portfolio vector represents the ratio of the investment in each asset/stock versus the total
budget invested, $w_t=[w_{0,t},w_{1,t} \dots w_{N,t}]$ s.t. $\sum_{i=1}^{N}w_{i,t} = 1$. $w_{0,t}$ is the fraction of cash we maintain at time $t$ and $w_{i,t}$ represents the fraction of investment on stock $i$ at time $t$. 

The portfolio value at time $t$ can then be written as $m_t=m_{t-1}v_t.w_{t-1}$, where $w_{t-1}$ is the portfolio weights chosen based on all information up to and including time $t-1$ and $m_{t-1}$ is the portfolio value at the previous time step.
Price changes over time reflect the profit or loss of the investment. To measure this, we use log returns. Log returns for period $t$ is the log of gross returns, which can be written as $R = \ln \frac{m_t}{m_{t-1}}$. Log returns provide a
symmetric way to calculate the future value of the portfolio unlike simple returns.
Finally, we have the risk measure, Value at Risk (VaR), defined as the maximum dollar amount expected to be lost over a given time horizon, at a pre-defined confidence level. 

In this work, we want to maximize the log returns of the portfolio while applying safety constraints such as limits on the Value at Risk(VaR) and other suitable conditions on the weights of the portfolio. The constrained optimization problem can be written as:
\begin{equation} \label{eq:1}
\begin{aligned}
& \max_{\pi}
& & \mathrm{R}(\pi) \\
& \text{s.t.}
& & Z(\pi) < \alpha. \\
\end{aligned}
\end{equation}
In the above Eqn., $\pi$ is a policy that updates the weights $w$ at each time step depending on the current state $x$ (which, for instance, can be a collection of relative price vectors $\{v_{u}\}_{u<t}$), $R(\pi)$ represents the cumulative sum of log returns, and $Z(\pi)$ represents the $m$ constraints that the policy needs to satisfy. To align with the prior literature, we will interchangeably use $a$ to represent the action (portfolio weights $w$) output by the policy $\pi$ when the context is clear.
We implement batch policy learning, which uses a pre-collected dataset $D = \{(x_i,a_i,x_i',r(x_i,a_i),g_{1:m}(x_i,a_i))\}_{i=1}^n$ generated from (potentially more than one) behavioural policies $\pi_{B}$, and as mentioned earlier, $m$ is the number of constraints imposed on the optimization variable.

\section{Approach}\label{sec:approach}

Our goal is to learn a policy $\pi$ from $D$ which satisfies the above constraints and maximizes the primary objective function.

\begin{algorithm}[H]
\caption{Constrained Portfolio Optimization}\label{alg:CPOT}
\hspace*{\algorithmicindent} \textbf{Input:} $D = \{(x_i,a_i,x_i',r_i,g_i)\}_{i=1}^n$\\
\hspace*{\algorithmicindent} \textbf{Output:} $\pi_{opt}$ 
\begin{algorithmic}[1]
\State Initialize $\lambda_0=(\lambda_0^1,\dots \lambda_0^{m+1}) s.t. \sum_{i=1}^{m+1} \lambda_0^i = B$
\For{\texttt{each Iteration t}}
\State \texttt{Learn policy},$\pi_t = FQI(D,\lambda_t)$
\State $\hat{R}(\pi_t) = FQE(\pi_t,r)$
\State $\hat{G}(\pi_t) = FQE(\pi_t,g)$
\State $\pi_{opt} = \frac{1}{t}\sum_{i=1}^t \pi_i$
\State $\hat{R}(\pi_t) = \frac{1}{t}\sum_{i=1}^t \hat{R}(\pi_i), \hat{G}(\pi_t) = \frac{1}{t}\sum_{i=1}^t \hat{G}(\pi_i)$
\State $\hat{\lambda}_t = \frac{1}{t}\sum_{i=1}^t \hat{\lambda}_i$
\State $L_{min} = \min_{\lambda}L(\pi_{opt},\lambda)$
\State $L_{max} = L(FQI(\hat{\lambda}_t),\hat{\lambda}_t)$
\If{$L_{max}-L_{min} < \omega$} 
\State{return} $\pi_{opt}$ 
\EndIf
\State $\lambda_{t+1}[i] = B \frac{\lambda_t[i]e^{-\eta grad[i]}}{\sum_j \lambda_t[j]e^{-\eta grad[j]}}$ where $grad=[(\hat{G}(\pi_t)-\tau)^T]$
\EndFor
\end{algorithmic}
\end{algorithm}

The Lagrangian for the above optimization Eqn.(\ref{eq:1}) can be written as $L(\pi,\lambda) = R(\pi) + \lambda^T (Z(\pi)-\alpha)$. Assuming strong duality, we can write the above problem as $\min_{\pi \in \Pi}\max_{\lambda} L(\pi,\lambda)$. This min max form reminds us of a two-player zero-sum game, where the first $\pi$-player tries to minimize $L(\pi,\lambda)$ for the current $\lambda$, and the second $\lambda$-player responds with a vector $\lambda$ (this is a standard interpretation of the reformulation). Under additional assumptions, it can be shown that an equilibrium can be reached by players playing in turn against each other over several rounds. Thus, after several times of performing such back and forth plays can lead us to a $\pi$, which is a candidate solution to our original problem. Therefore, to find such a policy, we use a no-regret learning algorithm to play the game as the second player ($\lambda$-player) repeatedly against a best response player ($\pi$-player) similar to \cite{freund1999adaptive}. When played this way, one can show that the the average of the actions played can converge to the solution of the game.

For best response, we consider mixed policies which are distributions over many deterministic policies, which we denote as $C(\Pi)$, and where $\Pi$ is the  set of  all deterministic policies. A randomized policy $pi$ from $C(\Pi)$ can be denoted as $\pi =\sum_{t=1}^{T} \eta_t \pi_t$ where $\pi_t \in \Pi$ and $\sum_{t=1}^T \eta_t =1$. Executing a mixed $\pi$ consists of first sampling one policy
$\pi_t$ from $\pi_{1:T}$ according to distribution $\eta_{1:T}$ , and then executing $\pi_t$.

As shown in Algorithm~\ref{alg:CPOT}, we use a no-regret algorithm for the $\lambda$-player and a best response strategy for the $\pi$-player. At each time $t=1 \dots T$, the learner makes a decision $\lambda_t \in \Lambda$ such that the regret, $Regret_t = \sum_t L(\pi_t,\lambda_t) - max_{\lambda} \sum_t L(\pi_t,\lambda_t)$, is minimized. For the learner to be no-regret, we require that $Regret_t = o(T)$. We use a standard batch reinforcement learning method to learn a policy that is $\textit{best-response}(\lambda_t) = \argmin_{\pi \in \Pi} L(\pi,\lambda_t) = \argmin_{\pi \in \Pi} R(\pi) + \lambda^T (Z(\pi)-\alpha)$.

Algorithm~\ref{alg:CPOT} outlines our approach of using a no-regret online learning algorithm and a batch policy optimization routine for $\lambda$ and $\pi$ respectively. For each iteration t, the $\pi-player$ runs the best response for a given $\lambda_t$. Then, we compute bounds on the objective, namely $L_{min}$ and $L_{max}$. This allows us to compute the primal-dual gap $L_{max} - L_{min}$. The game terminates when this gap is below a pre-specified threshold $\omega$.

\subsection{Player 1 - Policy Learning via Best Response}

For the $\pi$-player, we use a FQI (Fitted Q Iteration) as the best-response strategy given a fixed $\lambda$. The FQI algorithm is a batch mode reinforcement learning algorithm that yields an approximation of the Q-function corresponding to an infinite horizon optimal control problem
with discounted rewards in an iterative fashion.

\begin{algorithm}
\caption{FQI}\label{alg:FQI}
\hspace*{\algorithmicindent} \textbf{Input:} $D = \{(x_i,a_i,x_i',r_i,g_i)\}_{i=1}^n$, $\lambda$\\
\hspace*{\algorithmicindent} \textbf{Output:} $\pi$ 
\begin{algorithmic}[1]
\State Initialize $Q_0$ randomly
\For{\texttt{k=1\dots K}}
\State $ \text{Compute }y_i=(r_i+\lambda^T g_i)+\gamma\max_a Q_{k-1}(x_i',a) \text{ to generate dataset }$\par  \hskip\algorithmicindent $D_k=\{(x_i,a_i),y_i\}_{i=1}^n$
\State \text{Build a regression model $f(x_i,a_i)$ to solve for } \par
 \hskip\algorithmicindent $Q_k=\argmin_{f}\frac{1}{n}\sum_{i=1}^n (f(x_i,a_i)-y_i)^2$
\EndFor
\State \textbf{return} $\pi=\argmax_a Q_K(., a)$
\end{algorithmic}
\end{algorithm}

At the start of the FQI (see Algorithm~\ref{alg:FQI}), we randomly initialize $Q_0$. For each iteration $k=1 \dots K$, we build a new training dataset $\tilde{D}_k=\{(x_i,a_i),y_i\}_{i=1}^n$, which is a subset of the original dataset D and $y_i=(r_i+\lambda^T g_i)+\gamma\max_a Q_{k-1}(x_i',a)$. The data is shuffled so that the mini-batches used for training don't have highly correlated data. The batches of data are then used to learn the action-value function $Q_k$ using a supervised learning approach (i.e., by minimizing a loss that is a sum of individual observation specific losses). A stochastic gradient approach is used to obtain the action corresponding to a minimum Q-value at state $x_i'$. Once we calculate the target variable, we use a neural network as a function approximator to build a supervised regression model to solve for $Q_k$. This function can be written as $Q_k=\argmin_{f}\frac{1}{n}\sum_{i=1}^n (f(x_i,a_i)-y_i)^2$. Once the training is complete, we obtain the policy $\pi=\argmax_a Q_K(., a)$.

Note that FQI is not the only choice here. There are other best-response algorithms such as tree based methods \cite{ernst2005tree} and Least-Squares Policy Iteration \cite{lagoudakis2003least} that can also be used instead.

\subsection{Player 2 - No-regret Online Learning}
Once the final policy is obtained, we evaluate if the policy follows the objectives and constraints. We need to do this with having access only to the data $D$. This is called the \emph{off-policy policy evaluation} (OPE) problem. Few approaches used previously to solve this problem are: Importance Sampling (IS), Direct methods such as More Robust Doubly-Robust
(MRDR), Hybrid methods such as Weighted Doubly-Robust (WDR) etc. \cite{voloshin2019empirical}. We will focus on a technique called Fitted Q Evaluation (FQE), which was introduced in \cite{le2019batch}. It is a model free technique and is similar to the FQI technique, see Algorithm~\ref{alg:FQE}. The key difference is that the $\min$ operator in $y_i$ is replaced by $Q_{k-1}(x_i',\pi(x_i'))$. 

\begin{algorithm}[H]
\caption{FQE}\label{alg:FQE}
\hspace*{\algorithmicindent} \textbf{Input:} $\pi, e\in(r,g)$, $D = \{(x_i,a_i,x_i',r_i,g_i)\}_{i=1}^n$\\
\hspace*{\algorithmicindent} \textbf{Output:} $\hat{R}(\pi)$
\begin{algorithmic}[1]
\State Initialize $Q_0$ randomly
\For{\texttt{k=1\dots K}}
\State $ \text{Compute }y_i=e_i+\gamma Q_{k-1}(x_i',\pi(x_i')) \text{ to generate training dataset }$\par  \hskip\algorithmicindent $D_k=\{(x_i,a_i),y_i\}_{i=1}^n$
\State \text{Build a regression model $f(x_i,a_i)$ to solve for } \par
 \hskip\algorithmicindent $Q_k=\argmin_{f}\frac{1}{n}\sum_{i=1}^n (f(x_i,a_i)-y_i)^2$
\EndFor
\State \textbf{return} $\hat{R}(\pi) = Q_K(x,\pi(x))$
\end{algorithmic}
\end{algorithm}

Once the off-policy estimates of the objective and the constraints (the left hand side) of the current policy are computed, these can be used in two ways: (a) to estimate the primal-dual gap and terminate the two -player game, and (b) to update the current iterate of the $\lambda$-player (see Algorithm~\ref{alg:CPOT}).

Some choices for the latter are: Online Gradient Descent(OGD) \cite{zinkevich2003online} and Exponential Gradient (EG) \cite{kivinen1997exponentiated} methods, among others. Algorithm~\ref{alg:CPOT} uses Exponential Gradient, and since gradient-based algorithms typically require bounded $\lambda$ \cite{shalev2011online}, we force $\norm{\lambda}_1 < B$ using a pre-specified hyperparameter $B$. While solving Eqn.(\ref{eq:1}) requires setting $B=\infty$, we will set it to a finite value as needed for Algorithm~\ref{alg:CPOT}.

\section{Empirical Analysis}\label{sec:experiments}

We perform the following empirical evaluations: (i) we compare different off-policy evaluation techniques to assess their impact on the portfolio optimisation problem; (ii) we compare the performance of the final policy obtained using Algorithm~\ref{alg:CPOT} with baselines such as the Uniform Constant Rebalanced Portfolio (CRP), where the investment is equally distributed, and the pre-specified expert behaviour policy; and (iii) we study how the performance of the computed policy changes with portfolio size.

\subsection{Environment \& Data Collection}
The portfolio environment simulator is configured with the stock price data from $2012-08-13$ to $2017-08-11$. We generate the behavioural data by training an A2C model~\cite{mnih2016asynchronous} and collecting the episodes in the format $\{x_t,a_t,x_{t+1},r(x_t,a_t),VaR(x_t,a_t)\}$, where $x_t$ is the rolling window of relative prices $(v_t,...,v_{t-4})$, $a_t$ is nothing but the portfolio weight vector $w_t$, $r(x_t,a_t)$ is the  
log reward, and $VaR(x_t,a_t)$ is the constraint we intend to impose on the optimization problem. The dataset (and thus the behavior policy) need not abide by this constraint.

The specific optimization problem we solve is as below:
\begin{equation} \label{eq:3}
\begin{aligned}
& \underset{\pi}{\text{maximize}}
& & \mathrm{R}(\pi) \\
& \text{subject to}
& & VaR(x_t,\pi(x_t)) < 0.05 \forall t \in 1,...,T, \textrm{ and}\\
& & & 0.2 < a_i < 0.6\ \; \forall i \in 1,...,N, \textrm{ and}\\
& & & a_0 > 0.
\end{aligned}
\end{equation}

The constraints we are imposing are on the risk measure, Value at Risk (VaR) which is computed after every action. The first safety constraint we set is to ensure the Value at Risk is less than 5\% of our investment. Note that Value at Risk is a measure of the risk of loss for investments. The experiment is configured such that, with 95\% probability, one would lose at the most $5\%$ of their investment. For simplicity, we set the initial investment to $\$1$. So, the Value at Risk should not exceed $\$0.05$ at any given point in time. The second constraint is on the weights of the portfolio. This is to ensure that we don't invest all our money in a single stock. This is a simplistic constraint motivated by the observation that  traditionally, portfolio managers prefer to diversify their portfolio by investing in different sectors. We aim to replicate a similar effect using the second constraint. Finally, we note that the above optimization model is stylistic and does not capture all the real world details that is needed, but serves as a prototype for evaluating the effectiveness of off-policy optimization, which is the main goal of this work. An implementation of this experiment is made available on Github~\footnote{Code for reproducing the experimental results: \url{https://github.com/NymishaBandi/constrained-batch-policy-learning}}.

\subsection{Experimental Results on a Portfolio of Size 5}
Our first set of results are using a portfolio size of 5 stocks. Here we intend to study how different OPE techniques perform in different settings. We start with a portfolio of size 5 and measure portfolio performance using different OPE methods. We also vary the dataset size ($N$). It seems natural that having more data to train with will give a better policy. Our observations are consistent with that. The policy that performed the best is the one that was trained with the larger data size and used Fitted Q Evaluation as the OPE technique. Table~\ref{tab:p5} shows the overall results for the combination of different OPE techniques and dataset size for a setup with 5 stocks in the portfolio. 

\begin{figure}[H]
\begin{tabular}{cc}
  \includegraphics[width=65mm]{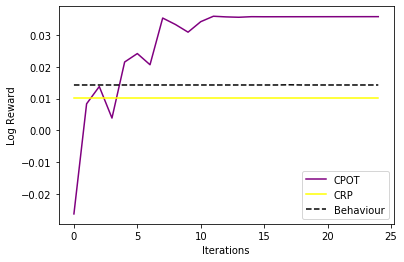} &   \includegraphics[width=65mm]{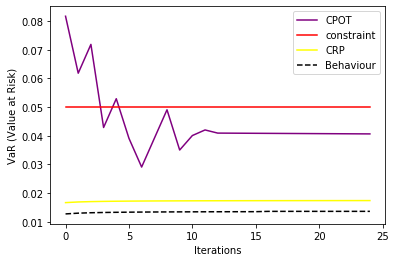} \\
\end{tabular}
\caption{Portfolio results for portfolio size 5. Comparing our algorithm with other RL algorithms without constraints (left) Improvement of the objective during the training. (right) Constraint satisfaction.\label{fig:p5}}
\end{figure}

The results for the above discussed settings that include the  CRP benchmark and the behaviour policy benchmark are shown in Fig.~\ref{fig:p5}. We can observe that the proposed constrained portfolio optimization method performs better than these benchmarks (the $x$-axis represents the number of iterations of Algorithm~\ref{alg:CPOT}). For instance, we note that our algorithm performs better than the behaviour policy, using whose data it learnt from.

\begin{table}[H]
\begin{tabular}{ c c c c c }
 \hline
 \multicolumn{5}{c}{Portfolio size $= 5$} \\
 \hline
 \multicolumn{3}{c}{$N=1028$}& \multicolumn{2}{c}{$N=2056$}\\
 \hline
 OPE &  Objective & Constraint &  Objective & Constraint\\
 \hline
 IS   &1.23E-3 &1.25E-1&6.29E-3 &9.21E-2  \\
 DR&  6.37E-3  & 6.75E-2&1.58E-2  & 1.21E-3 \\
 FQE & \textbf{1.92E-2} & \textbf{2.34E-1} & \textbf{3.57E-2} & \textbf{3.93E-3}\\
 \hline
\end{tabular}
\caption{Off-Policy Evaluation techniques for different Dataset size (N) - Comparing different OPE techniques w.r.t. main objective and constraints for portfolio size of $5$ when the dataset size is $N=1028$ and $2056$.\label{tab:p5}}
\end{table}

Figure~\ref{fig:p5} (right) shows the Value at Risk comparison for different methods. we observe that the benchmarks have a very low tolerance for risk, and give low returns accordingly. On the other hand, the policy obtained by Algorithm~\ref{alg:CPOT} starts off being highly risk-seeking, and eventually reaches a VaR that is less than the desired value.
Figure~\ref{fig:p5} (left) shows the objective value, which was the cumulative reward during the period of transactions. When we compare our results with the CRP benchmark and the behaviour policy, it is clear that we obtain a higher cumulative reward. It is to be noted that the benchmarks don't consider any constraints, unlike our algorithm.

\subsection{Experimental Results on a Portfolio of Size 10}
We perform a similar set of experiments as above with 10 stocks in the portfolio. Here, we have also updated the risk constraint so as to be more risk averse. The VaR is reduced to $0.035$. This should ideally give us a policy that is risk averse. The results obtained from the other policy evaluation techniques, namely Importance sampling (IS) and Doubly Robust (DR) methods, again turn out to be inferior to FQE. Further, they don't give us a policy that can obey the VaR constraint. On the other hand, FQE gives us better results (both the objective value and the VaR value) as compared to IS and DR (see Table~\ref{tab:p10}).

\begin{table}[H]
\begin{tabular}{ c c c c c }
 \hline
 \multicolumn{5}{c}{Portfolio size $= 10$} \\
 \hline
 \multicolumn{3}{c}{$N = 1028$}& \multicolumn{2}{c}{$N = 2056$}\\
 \hline
 OPE &  Objective & Constraint &  Objective & Constraint\\
 \hline
 IS   &6.34E-4 & 2.07E-1&1.06E-3 & 1.26E-1\\
 DR& 3.54E-3 & 7.92E-2&1.31E-2 & 4.61E-2 \\
 FQE & \textbf{2.22E-2} & \textbf{3.52E-2}& \textbf{4.97E-2} & \textbf{2.53E-2}\\
 \hline
\end{tabular}
\caption{Off-Policy Evaluation techniques for different Dataset size (N) - Comparing different OPE techniques w.r.t. the objective and constraint value (left hand side of the VaR constraint) for portfolio size of 10, when the dataset size is $N=1028$ and $2056$.\label{tab:p10}}
\end{table}

\begin{figure}[H]
\begin{tabular}{cc}
  \includegraphics[width=65mm]{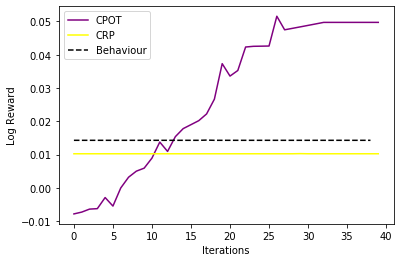} &   \includegraphics[width=65mm]{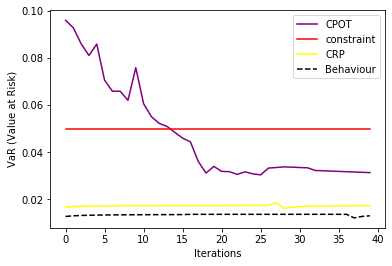} \\
\end{tabular}
\caption{Portfolio results for portfolio size 10. Comparing our algorithm with benchmarks: (left) Improvement of the objective during policy learning. (right) Constraint satisfaction.\label{fig:p10}}
\end{figure}

It is evident from the performance curves in Figure~\ref{fig:p10} that portfolio optimization for larger portfolio size takes longer to converge to an suitable policy. By changing the constraints in the optimisation problem, we observe that we can seamlessly get a risk averse or a risk seeking strategy as desired.

\section{Conclusion}
In this work, we developed an off-policy optimization method for portfolio allocation under constraints, and demonstrated its efficacy when compared to popular benchmarks under different settings. We also studied the performance of different Off-policy evaluation techniques and concluded that Fitted Q Evaluation seems to work the best for the portfolio allocation problem. Further investigation of these conclusions is needed when the optimization problem is made more realistic. The fact that we can use a single (behavior policy driven) data set to obtain multiple policies that are tailored to various scenarios makes our approach a promising candidate in developing good policies offline. In our experiments, we obtained a policy that is highly risk seeking and another policy that is risk averse using the same data set.

\bibliographystyle{plainnat}
\bibliography{portfolio}

\end{document}